\documentclass[letterpaper, 10 pt, conference]{ieeeconf}  

\IEEEoverridecommandlockouts                              

\usepackage{amsmath}
\usepackage{algorithm}
\usepackage[noend]{algpseudocode}
\usepackage{graphicx}
\usepackage{makecell}
\usepackage{booktabs}
\usepackage{subfig}
\usepackage{xcolor}
\usepackage{hyperref}
\let\labelindent\relax
\usepackage{enumitem}
\newcommand{\norm}[1]{\left\lVert#1\right\rVert}
\makeatletter
\def\BState{\State\hskip-\ALG@thistlm}
\makeatother
\DeclareMathOperator*{\argmin}{arg\,min}
\title{\LARGE \bf
Autonomous Tissue Manipulation via Surgical Robot Using \\ Learning Based Model Predictive Control
}
\author{Changyeob Shin$^{1}$, Peter Walker Ferguson$^{1}$, Sahba Aghajani Pedram$^{1}$, Ji Ma$^{1}$,\\ Erik P. Dutson$^{2,3}$, and Jacob Rosen$^{1,2,3}$
\thanks{$^{1}$Bionics Lab, Department of Mechanical and Aerospace Engineering, University of California, Los Angeles, Los Angeles, CA, USA 90095 (http://bionics.seas.ucla.edu)
}%
\thanks{$^{2}$Department of Surgery, David Geffen School
of Medicine, University of California, Los Angeles, Los Angeles, CA, USA 90095}%
\thanks{$^{3}$Center for Advanced Surgical and Interventional Technology (CASIT), University of California, Los Angeles, Los Angeles, CA, USA 90095}
\thanks{Email: \small  \{shinhujune, pwferguson, sahbaap, jima\}@ucla.edu,      EDutson@mednet.ucla.edu, jacobrosen@ucla.edu}
}

\begin{document}

\maketitle
\thispagestyle{empty}
\pagestyle{empty}

\begin{abstract}

Tissue manipulation is a frequently used fundamental subtask of any surgical procedures, and in some cases it may require the involvement of a surgeon's assistant. The complex dynamics of soft tissue as an unstructured environment is one of the main challenges in any attempt to automate the manipulation of it via a surgical robotic system. Two AI learning based model predictive control algorithms using vision strategies are proposed and studied: (1) reinforcement learning and (2) learning from demonstration. Comparison of the performance of these AI algorithms in a simulation setting indicated that the learning from demonstration algorithm can boost the learning policy by initializing the predicted dynamics with given demonstrations. Furthermore, the learning from demonstration algorithm is implemented on a Raven IV surgical robotic system and successfully demonstrated feasibility of the proposed algorithm using an experimental approach. This study is part of a profound vision in which the role of a surgeon will be redefined as a pure decision maker whereas the vast majority of the manipulation will be conducted autonomously by a surgical robotic system.
A supplementary video can be found at:  \textit{http://bionics.seas.ucla.edu/research/surgeryproject17.html}

\end{abstract}

\section{INTRODUCTION}
Automation in surgical robotics is part of a vision that will redefine the role of the surgeon in the operating room. It will shift the surgeons toward the decision making role while the vast majority of the manipulations will be conducted via a surgical robot. As part of this vision, research is directed at automating subtasks that serve as building blocks of many of the surgical procedures such as suturing \cite{pedram2017autonomous, dehghani2018automation,leonard2014smart}, tumor resection \cite{mckinley2016interchangeable}, bone cutting \cite{osa2014autonomous}, and drilling \cite{coulson2008autonomous}. Among the many surgical subtasks, tissue manipulation is one of the tasks that is most frequently performed. More specifically, when a surgeon wants to connect two different tissues or close an incision, both sides of the tissue should be placed with respect to each other in a way that enables homogeneous suture distance for improved healing \cite{waninger1992influence}. However, tissue manipulation presents a complex dynamics and hence is particularly challenging to automate given the lack of a model which predicts its behavior \cite{famaey2008soft}. Furthermore, indirect manipulation of interest points on the tissue makes it more difficult. Tissue manipulation falls under the broader research problem of deformable object manipulation.\\
\begin{figure}[!t]
\centering
\includegraphics[width=\linewidth]{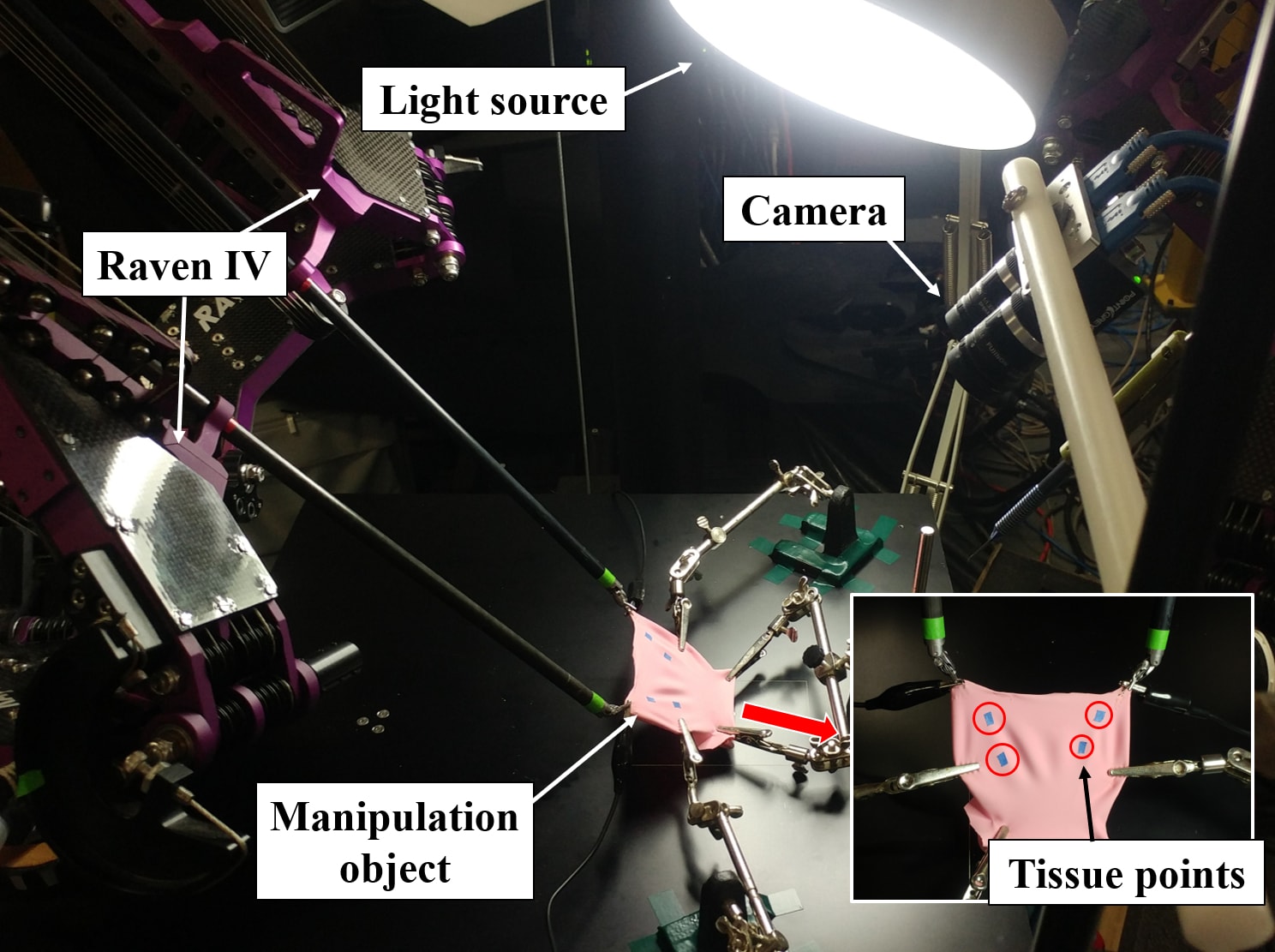}
\caption{Tissue manipulation experiment environment with the Raven IV surgical robotic system.}
\label{RobotExp}
\end{figure}
\indent There are in general two methods to approach the problem of tissue manipulation, namely model-based and model-free control. For model-based control method, a control law was suggested that could position a deformable object based on a spring-mass model and uncertainty \cite{hirai2000indirect}. In another study, a nonlinear finite element model was used to estimate the motion of soft tissue and parameters are updated using the difference between estimation and actual data \cite{boonvisut2013estimation}. In \cite{wada2001robust}, a PID controller was used with a model of a deformable object. The model-based manipulation of deformable objects is well summarized in \cite{jimenez2012survey}. For the model-free method, real time optimization framework utilizing rank-one Jacobian update with vision feedback has been used for manipulating a kidney \cite{alambeigi2018toward}, a deformable phantom tissue \cite{alambeigi2019autonomous},  and soft objects \cite{navarro2016automatic}. Furthermore, this model-free method has been expanded to manipulate a compliant object under unknown internal and external disturbances \cite{alambeigi2018robust}. A learned variable impedance control that trades off between force and position trajectories extracted from demonstrations is proposed for deformable objects manipulation \cite{lee2015learning}. In another study, linear actuators were controlled to apply external force to soft tissue to position a target feature while a needle is injected \cite{mallapragada2009robot}. Lastly, robotic manipulation and grasping of deformable objects are comprehensively covered in \cite{khalil2010dexterous}.

The reported research focuses on the task of manipulating tissue to place specified points on the tissue (tissue points) at desired positions in the image frame as described in Fig. \ref{RobotExp}. In operating rooms, tissue points can be selected as tissue features identifiable via recognition of common patterns on the tissue.  However, as part of this reported research, colored markers are used for robust tracking of the points with a vision algorithm. In the task, the tissue points are simultaneously indirectly manipulated by the robot arms grasping the tissue at manipulation points. This task is complicated by the complex dynamics between the motions of the robot arms and tissue points. This research effort proposes a learning-based model predictive control (MPC) framework to solve the dynamics and manipulate the soft tissue. Two learning approaches are compared. One is a reinforcement learning (RL) method where the robot learns the dynamics of the tissue after exploring by itself. The other is a learning from demonstration framework (LfD) that initializes the dynamics of the tissue by studying human expert demonstrations. 

Compared to previous model-based approaches that require a model for each manipulated object, the algorithms proposed in this study use a simple neural network to learn the dynamics in image space by exploiting the universality of neural network with hidden layers to describe any function. This provides flexibility of the algorithm to be applicable to any object, even those with different physical properties. As opposed to the reviewed model-free approaches using linearization, the proposed algorithms directly learn nonlinear dynamics of tissue in image space and control the robot with nonlinear optimization. This allows avoidance of local optima that may occur due to the physical constraints of the environment, by predicting future steps from learned dynamics. Moreover, the proposed LfD algorithm provides a framework to incorporate human demonstrations which leads to initialization of tissue dynamics and great controller performance even in the initial learning phase. The demonstrations can be easily acquired by recording the scene of robotic tissue manipulation in teleoperation mode.
\section{METHODS}
\subsection{Algorithms}
RL has shown great success in many applications where learning missing pieces, e.g. dynamics, of a task is necessary to find an optimal policy \cite{mnih2015human, silver2016mastering, kober2013reinforcement}. RL is a technique used by artificial agents or robots to learn strategies to optimize expected cumulative reward by collecting data through trial-and-error. As opposed to model-free RL, model-based RL is used in this work because of its high sample efficiency \cite{chebotar2017combining} which is desirable for robotic applications where collecting data with physical systems is expensive. Model-based RL updates a dynamics function with data samples collected by trial and error and has an internal controller to calculate control inputs. It optimizes a reward or cost function by applying the learned dynamics to the internal controller. In this work, model-based RL with internal MPC is used because it has been shown to successfully control robotic systems for a variety of tasks. An under-actuated legged robot was controlled in image space \cite{DBLP:journals/corr/abs-1711-05253}. An inverted pendulum was controlled in image space using an MPC paired with a learned deep dynamical model that predicts future images of the system \cite{wahlstrom2015pixels}.

\subsubsection{Assumptions}
We developed all algorithms based on the following assumptions:
\begin{itemize}[wide=0pt]
\item Vision feedback of robot and tissue features is always available. The robot and tissue features are never occluded.
\item The task begins after the robot grasps the manipulation points, and there is no slip between the grippers and the tissue.
\end{itemize}

\subsubsection{Model Predictive Control}
MPC is a control scheme that predicts future states by forward propagation using the current states, inputs, and dynamics equation in order to output the set of future inputs that result in optimal costs \cite{speyer2010primer}. The MPC has proven its ability to control complex mechanical systems \cite{eren2017model}. The MPC for tissue manipulation is formulated in this work with the following equations:
\begin{equation}
\begin{split}
\argmin_{\{u_t,\cdots,u_{t+h}\}}&{ \norm{\vec{p}_{t+h+1}^{\,T,des}-\vec{p}_{t+h+1}^{\,T,curr}}_2^2} \\
s.t. \quad &\vec{v}_{t+h}^{\,T}=f(\vec{p}_{t+h}^{\,T}, \vec{p}_{t+h}^{\,R}, \vec{u}_{t+h}^{R}) \\
&\vec{p}_{t+h+1}^{\,T}=\vec{p}_{t+h}^{\,T}+\int_{0}^{\Delta t} \vec{v}_{t+h}^{\,T} dt \\
&\vec{p}_{t+h+1}^{\,R}=\vec{p}_{t+h}^{\,R}+\int_{0}^{\Delta t} \vec{u}_{t+h}^{\,R} dt \\
& h = 0, \cdots, H-1 \\
\end{split}
\end{equation}
where 
superscript $T$ and $R$ are used to designate the tissue points and robot wrists. $u$ is an input specifying movement of the robot in image space, $\Delta t $ is control period, and $H$ is the maximum number of steps in the time horizon. $\vec{p}^{T}$ and $\vec{v}^{T}$ are position and velocity vectors defined in image space, of all tissue points, and are each $\in R^{2*\{\# \textit{ of tissue points}\}}$. In the same manner, $\vec{p}^{R} \in R^{2*\{\# \textit{ of robots}\}}$ and $\vec{v}^{R} \in R^{2*\{\# \textit{ of robots}\}}$. The cost function is formulated to reduce the Euclidean distance between the tissue points and desired points at a time instance. If the tissue points are close to their desired positions, it is not necessary to use all inputs in the input horizon. Thus, the optimal number of inputs in the input horizon is also found in equations Ò(1)Ó. As a result, the output from the MPC formulated in equations Ò(1)Ó is a set $\{u_t^*,\cdots,u_{t+h^*}^*\}$.

\subsubsection{Adaptive MPC}
Accurate modeling of dynamics is crucial for successful application of MPC. However, defining the dynamics of a complex system is challenging. In order to address this challenge, adaptive MPC that updates the dynamics using learning algorithms was suggested \cite{hedjar2013adaptive, chowdhary2013concurrent}. In this work, a neural network is used to find the dynamics for the MPC. The input vector for the dynamics neural network is a vector that is composed of positions of the robot wrists, positions of the tissue points, and the control inputs for the robots. The output is the velocities of the tissue points.

An optimal control sequence, which is an output from the equation Ò(1)Ó, can be obtained in two ways: optimization by error back propagation \cite{wahlstrom2015pixels} or generation of random input candidates \cite{DBLP:journals/corr/abs-1711-05253}. The research presented in this study uses the latter approach for calculating the optimal number of steps in the time horizon $h^*$ and the corresponding control sequence. After that, for each control period, the desired robot wrist positions in image space are updated with the first input in the optimal control sequence. The robot position is controlled to match the desired positions via visual servoing.

\begin{figure}[!t]
\centering
\vspace{+0.3cm}
\includegraphics[width=\linewidth]{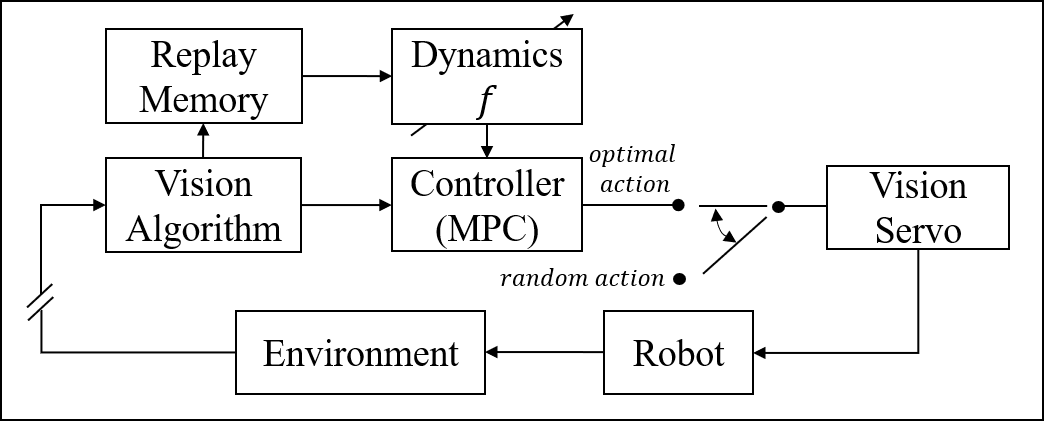}
\captionsetup{singlelinecheck = false}
\caption{Block diagram of Reinforcement Learning.}
\label{RL_Schematic}
\end{figure}
\begin{algorithm}[!tb]
\caption{Model-based Reinforcement Learning}\label{RLAl}
\begin{algorithmic}[1]
\State $\textit{initialize neural network variables}$
\While {action number $<$ exploration number}
\State $\textit{extract $p_t^T$ and $p_t^R$ in image}$
\If {$\textit{frame}_{curr} - \textit{frame}_{prev} > \Delta t$}
\State $\textit{with} \,\epsilon, \textit{choose optimal or random action}$
\State $\textit{update $p_{t+1}^{R,des}$ in image}$
\State $\textit{save experience to replay memory}$
\State $\textit{choose mini batch and train dynamics model}$
\EndIf
\State $\textit{visual servoing of robots}$
\EndWhile
\end{algorithmic}
\end{algorithm}

\subsubsection{Model-based Reinforcement Learning}
The reinforcement learning algorithm using model predictive control is shown in Fig. \ref{RL_Schematic} and summarized in Algorithm \ref{RLAl}.  After initialization of the neural network variables, a computer vision algorithm extracts the positions of the robot wrists and the tissue points from an image. $\epsilon$-greedy approach is used to force the robot to explore randomly at the beginning, but gradually optimize policy as dynamics are learned. For $\epsilon$-greedy behavior in RL, $ \epsilon $ is decreased from 1 to 0.1 linearly as a function of the number of robot actions taken. If $\epsilon$ is greater than a random number generated between 0 and 1, a random action is taken. Otherwise, the optimal action based on the MPC is taken. Each action set, $u \in \{[0,0], [1,0], [-1,0], [0,1], [0,-1]\}$ is multiplied by a scale factor, $step$, in image space that determines the step size of the robots in pixels. The actions in $u$, in order, correspond to the robot stopping, moving left, moving right, moving upward, and moving downward in image space. We have found that step size of the robots' movement and control period should be properly selected to learn meaningful information. After the action is determined, desired robot positions are updated and visual servo is performed to control the robots. In the next frame, the algorithm again obtains positions, and calculates velocities of the robot wrists and tissue points by taking the difference between previous and current positions. An experience set that contains the previous positions and velocities is saved to the replay memory. Training of the dynamics model starts after collecting more than a predefined number of experience sets. Random but fixed size sets of experience sets are selected from the replay memory and used to train the network. This process is repeated until the number of total actions reaches a predefined exploration number.  After this learning period, the robot always chooses the optimal action based on the learned dynamics. 

\begin{figure}[!t]
\centering
\vspace{+0.3cm}
\includegraphics[width=\linewidth]{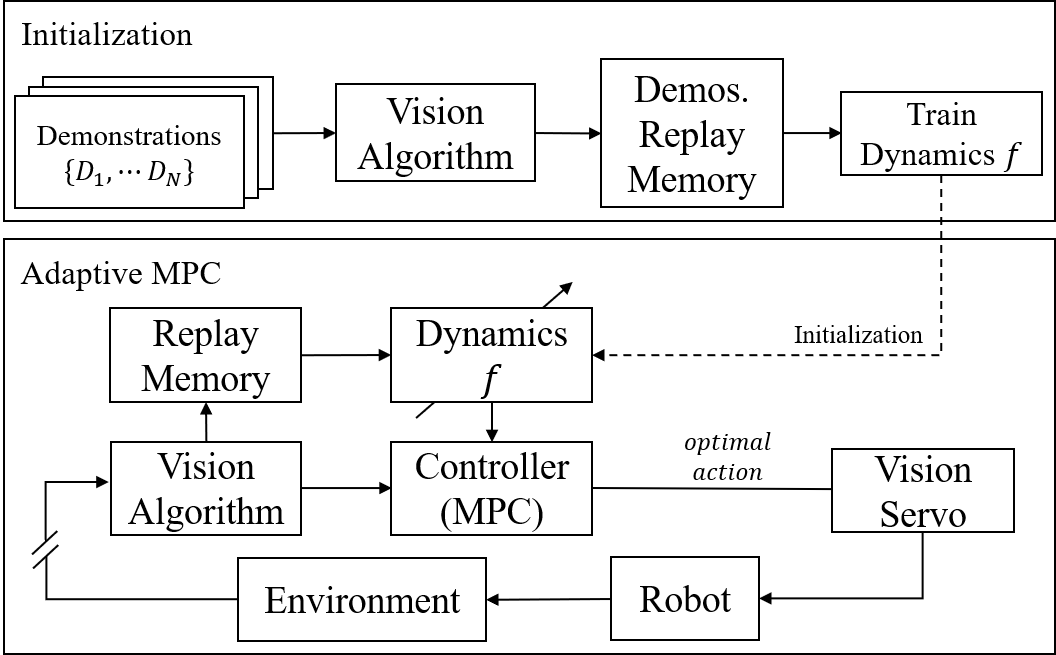}
\captionsetup{singlelinecheck = false}
\caption{Block diagram of Learning from Demonstration.}
\label{LfD}
\end{figure}
\begin{algorithm}[!t]
\caption{Learning from Demonstrations}\label{LfDAl}
\begin{algorithmic}[1]
\State $\textit{initialize neural network variables}$
\For {Number of frames in demonstrations}
\State $\textit{extract $p^T$ and $p^R$ in image}$
\State $\textit{save experience to replay memory}$ 
\EndFor
\State $\textit{train dynamics network with replay memory}$
\State $\textit{initialize dynamics of MPC with the trained network}$
\While {$error > threshold$}
\State $\textit{extract $p_t^T$ and $p_t^R$ in image}$
\If {$\textit{frame}_{curr} - \textit{frame}_{prev} > \Delta t$}
\State $\textit{choose optimal action}$
\State $\textit{update $p_{t+1}^{R,des}$ in image}$
\State $\textit{save experience to replay memory}$
\State $\textit{choose mini batch and train dynamics model}$
\EndIf
\State $\textit{visual servoing of robots}$
\EndWhile
\end{algorithmic}
\end{algorithm}

\subsubsection{Learning from Demonstrations}
Self exploration can be time-consuming and dangerous when the robot does not have any prior knowledge of the environment. As such, learning the dynamics from scratch using RL is inadvisable for tissue manipulation in clinical environments. However, if human experts such as surgeons can demonstrate the task to the robots, and if the robots can learn from these demonstrations, it would be safer. LfD is actively studied in the field of robotics because of its many strengths \cite{argall2009survey}. Furthermore, it has been shown that the learning process in model-based reinforcement learning can be accelerated from demonstrations \cite{schaal1997learning}. These demonstrations could be obtained, for tissue manipulation during surgery, by surgeons recording videos that capture the screen of the teleoperation console which is actively used for controlling surgical robots \cite{li2011maximizing}. Thus, an LfD algorithm that initializes the dynamics using experts' demonstrations is proposed in Fig. \ref{LfD}. We assume that demonstrations are in video formats that consist of a sequence of images and that the video captures teleoperation of surgical robots by human experts.

The LfD algorithm is described in Algorithm \ref{LfDAl}. In the first phase of the LfD algorithm, images from demonstrations are fed to the vision algorithm, and the positions and velocities of the robot wrists and tissue points are extracted. Experience sets are saved to the demonstration replay memory and the dynamics neural network is trained with this memory. After this training phase with demonstrations, the dynamics neural network in MPC is initialized with the trained network. This research found that if demonstrations for only one specific set of desired tissue point positions are given, robots can only properly locate the tissue points to desired positions slightly different than in the demonstrations. In order to reach significantly different sets of desired tissue point positions, exploration is necessary. However, if demonstrations encompass a wide range of desired positions and workspaces of the robots, the robots can finish tasks even without exploration.

\subsubsection{Computer Vision Algorithm} 
Positions of the robots and tissue features in image space are extracted through a computer vision algorithm. Robot wrist and tissue point positions are recognized by colored features installed at appropriate locations. The position of each component is calculated as the average of the contour point positions that can be obtained after color segmentation and morphological operations. This research used the OpenCV library for processing these operations \cite{opencv_library}. The tissue points are labeled based on the prior information of the configuration. The motions of the robot wrists are restricted to a two-dimensional square workspace to prevent occlusion of the tissue points. However, this workspace restriction limits the dexterity of the robots.
\subsubsection{Learning Algorithm Hyperparameters} 
The dynamics neural network was chosen to have two hidden layers with 12 elements each. Rectified Linear Unit (ReLU) is used for the activation function for both hidden layers. 
Four tissue points are used in this research effort, resulting in input and output vector sizes of 16 and 8 respectively. The method presented in this research effort can be easily scalable to different number of tissue points and robot arms. Four tissue points and two grasp points are chosen to demonstrate the viability of the proposed algorithm even when an exact solution does not exist because there is insufficient controllability. Weights of the networks are initialized with random numbers from a normal distribution. Learning rate was set to 0.01 and batch size to 200 for RL. Adam optimizer was used to train the neural networks \cite{kingma2014adam}. To stay within computational and physical limitations of the computer and robot, the control period, $\Delta t$, was set to 0.5 sec for both RL and LfD algorithms. $Step$ was set to 5 during the learning process in RL. For RL in simulation, the episode was reset every 1,000 actions and the robot was allowed to explore the state space until it reached 5,000 actions. 

\setlength{\textfloatsep}{5 pt}
\begin{figure}[!t]
\centering
\vspace*{0.15cm}
\includegraphics[width=\linewidth]{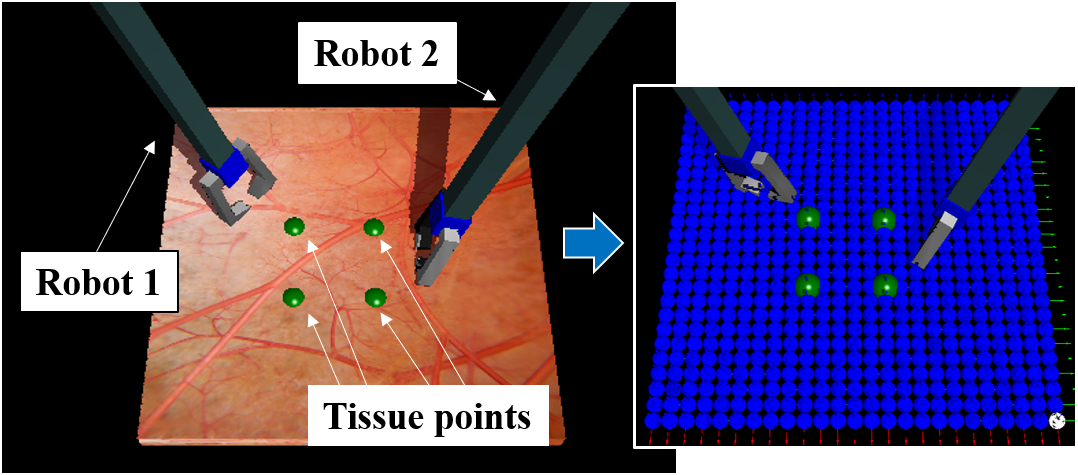}
\caption{Simulation environment and skeleton structure of soft tissue.}
\label{Sim}
\end{figure}
\begin{figure*}[!t]
\centering
\subfloat[]{\includegraphics[width=0.24\textwidth]{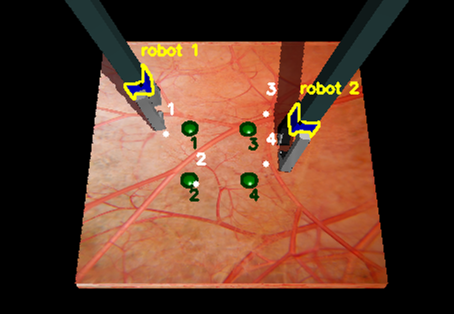}
\label{SimProcess:a}}
\hfil
\subfloat[]{\includegraphics[width=0.24\textwidth]{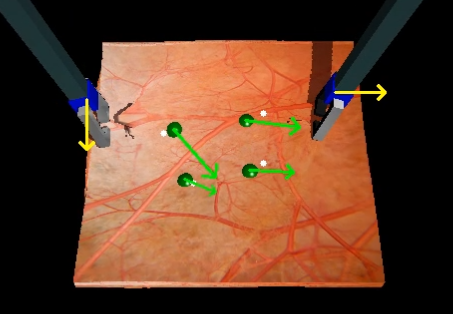}
\label{SimProcess:b}}
\hfil
\subfloat[]{\includegraphics[width=0.24\textwidth]{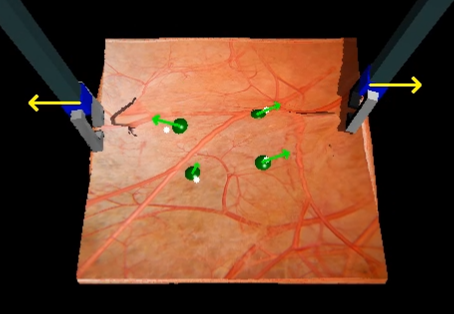}
\label{SimProcess:c}}
\hfil
\subfloat[]{\includegraphics[width=0.24\textwidth]{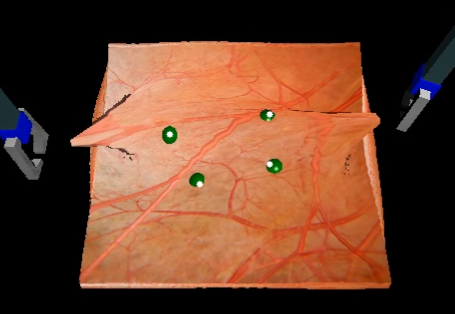}
\label{SimProcess:d}}
\caption{Illustration of sequence of simulation experiment. (a) Initial configuration with computer vision results marked as yellow (robot wrists), green (tissue points), and white (desired tissue point positions). (b) Learned dynamics when left robot moves downward and right robot moves right, yellow arrows scaled 10 times show movement of robot and green arrows scaled 30 times visualize predicted motion of tissue points by the robots' motions. (c) Learned dynamics when left robot moves left and right robot moves right. (d) Final configuration showing tissue points located at the desired positions.}
\label{SimProcess}
\end{figure*}

\subsection{Simulation}
To verify the performance of the controller, a tissue manipulation simulation was designed. Fig. \ref{Sim} demonstrates the environment of the simulation. We used CHAI3D open-platform simulation \cite{Conti03}. The GEL module is used to describe the motion of the soft tissue. The simulated tissue consists of a predefined number of spheres as illustrated in Fig. \ref{Sim}. The physical properties of the soft tissue can be set by mass, spring, and damper coefficients for the nodes that form the skeleton structure of the soft tissue, and these physical properties determine the tissue dynamics. The dynamics of the simulation update at a frequency of 1kHz. The movements of the tissue elements are determined by the library of GEL based on the given external forces. External attraction forces between two manipulation points and two robot grippers are generated proportional to the distance between them. The positions of elements on the boundary of the tissue were fixed for internal stability of the tissue. The two manipulation points, four tissue points, and the desired positions of the tissue points are predetermined at the beginning of the simulation. Green markers are attached to the tissue points and wrists of the robots are colored as blue. Human operators can demonstrate bimanual manipulation of the simulated tissue with two phantom omni \cite{silva2009phantom}.

\begin{figure}[!t]
\centering
\vspace{-0.5 cm}
\includegraphics[width=\linewidth]{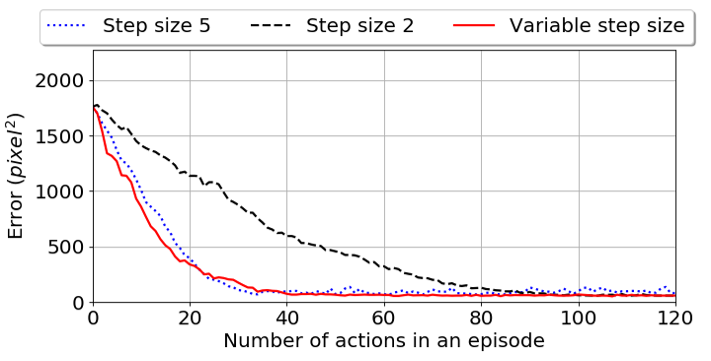}
\caption{Positioning error versus action number in simulation experiments with varied $step$ used in the fully trained controller from RL.}
\label{SimPosError}
\end{figure}

\subsection{Surgical Robot Experiment}
\subsubsection{Raven IV}
Experiments were performed with the Raven IV, an open platform for surgical robot research \cite{li2011maximizing}. The Raven IV possesses two pairs of cable-driven surgical robotic arms, each with 7-degrees-of- freedom (DoF) including the grippers. In the experiments conducted, only one pair of robot arms was used. A surgeon or separate surgical automation algorithm could independently perform other tasks, such as suturing, with the remaining two robot arms.

\subsubsection{Experiment Environment}
The environment of the experiment is shown in Fig. \ref{RobotExp}. The manipulation object is made of highly elastic colored latex, and is used to emulate tissue. Four clips are used to fix the object while the robot performs the task. Blue tape is attached to the manipulation object to represent the tissue points and facilitate tracking using the computer vision algorithm. Although a stereo camera (Blackfly-BFLY-U3-13S2C, Point Grey Research) is shown in Fig. \ref{RobotExp}, it is used in single camera mode. The original resolution of the camera is 1288x964 but it was reduced to 644x482 before processing the computer vision algorithm. The frame rate of the camera is 30 Hz. For robustness of the computer vision algorithm, a light source is installed.

\section{Results and Discussion}
\subsection{Simulation}
Both RL and LfD algorithms were implemented in the simulation and were evaluated. For the MPC, the maximum number of steps in the time horizon was set to 5, and 5,000 sets of robot action candidates were generated and evaluated during each control period.
\subsubsection{Effects of Step Size}
Fig. \ref{SimPosError} illustrates the positioning error of the tissue points in an episode with different step sizes. As is shown, the neural network functions and the fully trained controller from the RL algorithm successfully minimizes error regardless of step size. Steady state error is not zero because it is not always feasible to place the four tissue points at exactly their desired positions using only two robot arms.  As can be seen, depending on step size, the decay rate and steady state of the error vary. When the step size is large ($step$=5), error decreases quickly but steady state error oscillates because the appropriate step size near the destination is less than the predetermined step size. Alternatively, small step size ($step$=2) results in slow error decay rate but stable and smaller steady state error. Therefore, a variable step size is used. When error is greater than 150, a large step size is used to quickly decrease error. When error is between 150 and 70, a small step size is used to stabilize and reduce the steady state error. Below 70 error, $step$ is further reduced to unity (actions move the robot arms single pixels). Both RL and LfD algorithms use the variable step size. The sequence of the simulation is illustrated in Fig. \ref{SimProcess}. The initial configuration is shown in Fig. \ref{SimProcess:a}, with the desired tissue point positions labeled 1 through 4. As the simulation progresses in Figs. \ref{SimProcess}b-c, the robot actions (yellow arrows) are input to the learned dynamics to predict the motion of the tissue points (green arrows). Eventually, the error between the actual and desired tissue point positions is minimized as shown in Fig. \ref{SimProcess:d}, and the simulation ends. 

\begin{figure}[!t]
\centering
\vspace{-0.6 cm}
\includegraphics[width=\linewidth]{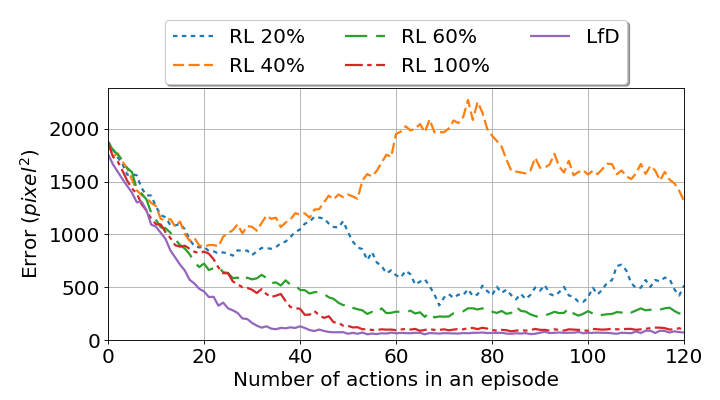}
\caption{Comparison of LfD and RL at multiple learning stages. RL applied its learned dynamics at each stage and fully exploited optimal actions.}
\label{RLvsLfD}
\end{figure}
\begin{figure*}[!t]
\centering
\subfloat[]{\includegraphics[width=0.24\textwidth]{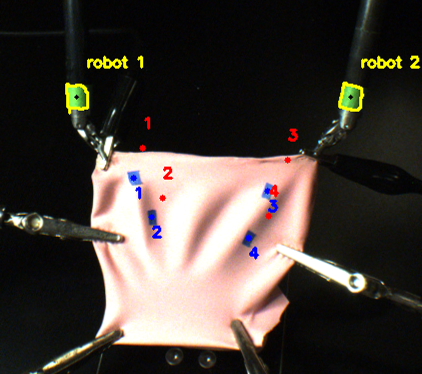}
\label{RobotExpRes:a}}
\hfil
\subfloat[]{\includegraphics[width=0.24\textwidth]{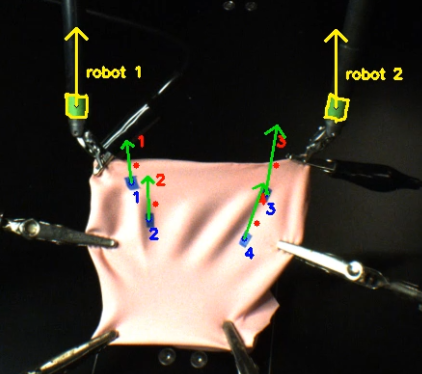}
\label{RobotExpRes:b}}
\hfil
\subfloat[]{\includegraphics[width=0.24\textwidth]{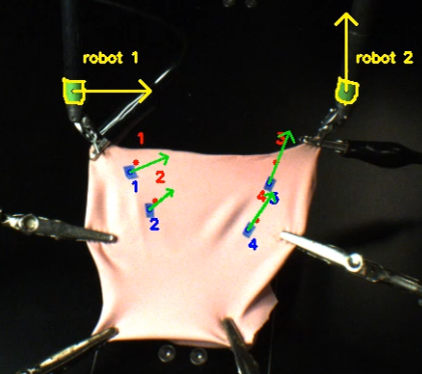}
\label{RobotExpRes:c}}
\hfil
\subfloat[]{\includegraphics[width=0.24\textwidth]{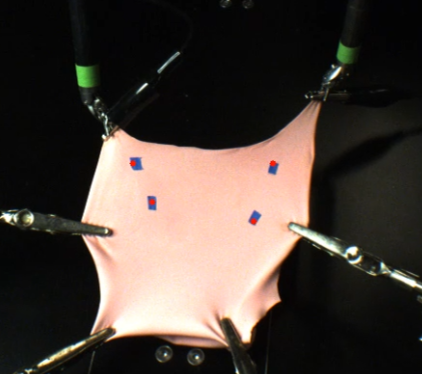}
\label{RobotExpRes:d}}
\caption{Illustration of the robot experiment sequentially from (a) to (d). (a) Shows the initial configuration and computer vision algorithm results, red dots are desired tissue points positions. (b) Illustrates learned dynamics when both robots move upward, yellow arrows scaled 10 times represent the motion of robots and green arrows scaled 10 times visualize predicted motions of the tissue points. (c) Shows learned dynamics when left robot moves right and right robot moves upward. (d) Final configuration of experiment when the task is finished.}
\label{RobotExpRes}
\end{figure*}
\subsubsection{Simulation RL vs. LfD}
To compare the RL and LfD algorithms in simulation, three demonstrations were collected from one expert and used to train the dynamics network. We compare the performance of the controller with LfD and at different stages of RL in Fig. \ref{RLvsLfD}. For this comparison, RL fully exploited optimal actions based on the current dynamics it has at each learning stage. As expected, RL does not perform well until it has been thoroughly trained. We also observe that controller performance does not necessarily improve during the process of learning until dynamics are well understood. This is evident in a comparison of RL 40\% and RL 20\% in Fig. \ref{RLvsLfD}. However, LfD is able to perform the task immediately after initialization when desired tissue point positions are not far from the ones in the demonstrations. It was found that if a single demonstration covers a wide range of the workspace, the single demonstration is sufficient for moving tissue points to a variety of desired positions. 

\subsection{Surgical Robot Experiment}
From the simulation results in the initial states of training the RL without LfD, it was judged that RL is potentially hazardous and too time consuming to apply to physical systems. However, based on the results of the simulation, it was observed that the initial policy from LfD is meaningful enough to perform the task on the Raven IV. For the MPC, the maximum number of steps in the time horizon was set to 12, and 10,000 sets of robot action candidates were generated and evaluated during each control period.

\begin{figure}[!t]
\vspace{-0.5 cm}
\centering
\includegraphics[width=\linewidth]{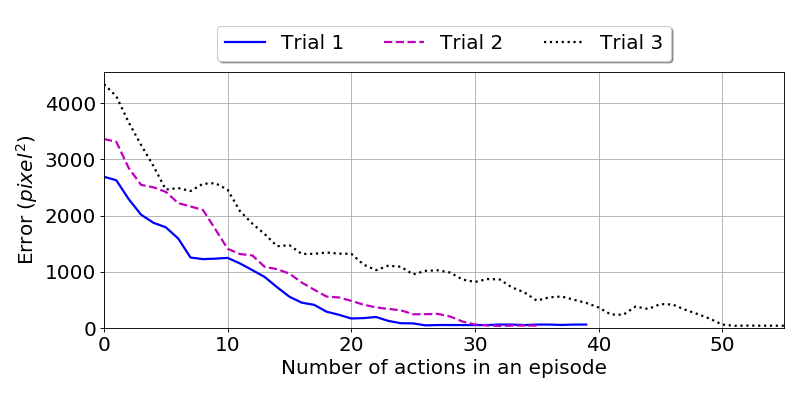}
\caption{Experiment results of LfD with Raven IV. The experiments were repeated three times on similar initial configurations.}
\label{RobotExpErrorPlot}
\end{figure}
Two demonstrations were collected by an operator controlling the Raven IV in teleoperation mode with camera feedback, and used to initialize the neural network. We repeated the robot experiment three times with similar initial configurations of manipulation points, initial tissue point positions, and desired tissue point positions. A sequence of captured images representative of these experiments is demonstrated in the Fig. \ref{RobotExpRes}. The operation of the robot was stopped when it placed the four tissue points to the desired tissue point positions with a total error of less than 50 pixels squared as shown in Fig. \ref{RobotExpErrorPlot}. Note that there are slight differences in initial errors among the experiments because initial tissue point positions cannot be replicated exactly. For these experiments, $step$ was varied by switching between 8 and 4. Fig. \ref{RobotExpRes:d} shows that the robot could successfully position the tissue points by stretching the object.
\section{CONCLUSIONS}
In this research effort, RL and LfD algorithms have been presented for automating the soft tissue manipulation task. Experiments on the simulation showed that both algorithms could accomplish the task. The results demonstrate that LfD boosts the learning process by dramatically reducing the amount of exploration required for the neural network to correctly learn the dynamics. The LfD algorithm was implemented on the Raven IV surgical robot, and the robot successfully placed the tissue points to their desired positions. This shows that the LfD algorithm can result in a good initial policy for accomplishing the task in real environments when relevant workspaces are covered in demonstrations. The authors believe that the capability of the LfD algorithm could be expanded significantly if given additional demonstration data that more fully captures the robot's workspace.

There are limitations to the proposed approach that will be addressed by future studies. In the simulation and experiments, the workspace of the robots was constrained to prevent visual occlusion. However, in order to utilize the full capabilities of robots, an algorithm that can avoid the occlusion may be developed. In addition, the algorithm should be expanded to three-dimensional tissue manipulation. Furthermore, the LfD framework presented in this work provides good initial dynamics based on the assumption that the relationship between the camera frame and environment frame is fixed. There should be further research to efficiently use the demonstration data provided in different camera frames than the task environment.

In conclusion, it is anticipated the future research along similar lines will eventually lead to the introduction of automation into surgery clinically in which subtasks of the surgical procedure will be fully automated. This approach is likely to unify across the field that will eventually lead to improved patents’ outcome.

\addtolength{\textheight}{-3.2cm}   

\bibliographystyle{ieeetr}
\bibliography{Bib/mybibfile.bib}

\begin{thebibliography}{10}

\bibitem{pedram2017autonomous}
S.~A. Pedram, P.~Ferguson, J.~Ma, E.~Dutson, and J.~Rosen, ``Autonomous
  suturing via surgical robot: An algorithm for optimal selection of needle
  diameter, shape, and path,'' in {\em Robotics and Automation (ICRA), 2017
  IEEE International Conference on}, pp.~2391--2398, IEEE, 2017.

\bibitem{dehghani2018automation}
H.~Dehghani, S.~Farritor, D.~Oleynikov, and B.~Terry, ``Automation of suturing
  path generation for da vinci-like surgical robotic systems,'' in {\em 2018
  Design of Medical Devices Conference}, pp.~V001T07A008--V001T07A008, American
  Society of Mechanical Engineers, 2018.

\bibitem{leonard2014smart}
S.~Leonard, K.~L. Wu, Y.~Kim, A.~Krieger, and P.~C. Kim, ``Smart tissue
  anastomosis robot (star): A vision-guided robotics system for laparoscopic
  suturing,'' {\em IEEE Transactions on Biomedical Engineering}, vol.~61,
  no.~4, pp.~1305--1317, 2014.

\bibitem{mckinley2016interchangeable}
S.~McKinley, A.~Garg, S.~Sen, D.~V. Gealy, J.~P. McKinley, Y.~Jen, M.~Guo,
  D.~Boyd, and K.~Goldberg, ``An interchangeable surgical instrument system
  with application to supervised automation of multilateral tumor resection.,''
  in {\em CASE}, pp.~821--826, 2016.

\bibitem{osa2014autonomous}
T.~Osa, C.~F. Abawi, N.~Sugita, H.~Chikuda, S.~Sugita, H.~Ito, T.~Moro,
  Y.~Takatori, S.~Tanaka, and M.~Mitsuishi, ``Autonomous penetration detection
  for bone cutting tool using demonstration-based learning,'' in {\em Robotics
  and Automation (ICRA), 2014 IEEE International Conference on}, pp.~290--296,
  IEEE, 2014.

\bibitem{coulson2008autonomous}
C.~Coulson, R.~Taylor, A.~Reid, M.~Griffiths, D.~Proops, and P.~Brett, ``An
  autonomous surgical robot for drilling a cochleostomy: preliminary porcine
  trial,'' {\em Clinical Otolaryngology}, vol.~33, no.~4, pp.~343--347, 2008.

\bibitem{waninger1992influence}
J.~Waninger, G.~W. Kauffmann, I.~A. Shah, and E.~H. Farthmann, ``Influence of
  the distance between interrupted sutures and the tension of sutures on the
  healing of experimental colonic anastomoses,'' {\em The American journal of
  surgery}, vol.~163, no.~3, pp.~319--323, 1992.

\bibitem{famaey2008soft}
N.~Famaey and J.~V. Sloten, ``Soft tissue modelling for applications in virtual
  surgery and surgical robotics,'' {\em Computer methods in biomechanics and
  biomedical engineering}, vol.~11, no.~4, pp.~351--366, 2008.

\bibitem{hirai2000indirect}
S.~Hirai and T.~Wada, ``Indirect simultaneous positioning of deformable objects
  with multi-pinching fingers based on an uncertain model,'' {\em Robotica},
  vol.~18, no.~1, pp.~3--11, 2000.

\bibitem{boonvisut2013estimation}
P.~Boonvisut and M.~C. {\c{C}}avu{\c{s}}o{\u{g}}lu, ``Estimation of soft tissue
  mechanical parameters from robotic manipulation data,'' {\em IEEE/ASME
  Transactions on Mechatronics}, vol.~18, no.~5, pp.~1602--1611, 2013.

\bibitem{wada2001robust}
T.~Wada, S.~Hirai, S.~Kawamura, and N.~Kamiji, ``Robust manipulation of
  deformable objects by a simple pid feedback,'' in {\em Robotics and
  Automation, 2001. Proceedings 2001 ICRA. IEEE International Conference on},
  vol.~1, pp.~85--90, IEEE, 2001.

\bibitem{jimenez2012survey}
P.~Jim{\'e}nez, ``Survey on model-based manipulation planning of deformable
  objects,'' {\em Robotics and computer-integrated manufacturing}, vol.~28,
  no.~2, pp.~154--163, 2012.

\bibitem{alambeigi2018toward}
F.~Alambeigi, Z.~Wang, Y.-h. Liu, R.~H. Taylor, and M.~Armand, ``Toward
  semi-autonomous cryoablation of kidney tumors via model-independent
  deformable tissue manipulation technique,'' {\em Annals of Biomedical
  Engineering}, pp.~1--13, 2018.

\bibitem{alambeigi2019autonomous}
F.~Alambeigi, Z.~Wang, R.~Hegeman, Y.-H. Liu, and M.~Armand, ``Autonomous
  data-driven manipulation of unknown anisotropic deformable tissues using
  unmodelled continuum manipulators,'' {\em IEEE Robotics and Automation
  Letters}, vol.~4, no.~2, pp.~254--261, 2019.

\bibitem{navarro2016automatic}
D.~Navarro-Alarcon, H.~M. Yip, Z.~Wang, Y.-H. Liu, F.~Zhong, T.~Zhang, and
  P.~Li, ``Automatic 3-d manipulation of soft objects by robotic arms with an
  adaptive deformation model,'' {\em IEEE Transactions on Robotics}, vol.~32,
  no.~2, pp.~429--441, 2016.

\bibitem{alambeigi2018robust}
F.~Alambeigi, Z.~Wang, R.~Hegeman, Y.-H. Liu, and M.~Armand, ``A robust
  data-driven approach for online learning and manipulation of unmodeled 3-d
  heterogeneous compliant objects,'' {\em IEEE Robotics and Automation
  Letters}, vol.~3, no.~4, pp.~4140--4147, 2018.

\bibitem{lee2015learning}
A.~X. Lee, H.~Lu, A.~Gupta, S.~Levine, and P.~Abbeel, ``Learning force-based
  manipulation of deformable objects from multiple demonstrations,'' in {\em
  Robotics and Automation (ICRA), 2015 IEEE International Conference on},
  pp.~177--184, IEEE, 2015.

\bibitem{mallapragada2009robot}
V.~G. Mallapragada, N.~Sarkar, and T.~K. Podder, ``Robot-assisted real-time
  tumor manipulation for breast biopsy,'' {\em IEEE Transactions on Robotics},
  vol.~25, no.~2, pp.~316--324, 2009.

\bibitem{khalil2010dexterous}
F.~F. Khalil and P.~Payeur, ``Dexterous robotic manipulation of deformable
  objects with multi-sensory feedback-a review,'' in {\em Robot Manipulators
  Trends and Development}, InTech, 2010.

\bibitem{mnih2015human}
V.~Mnih, K.~Kavukcuoglu, D.~Silver, A.~A. Rusu, J.~Veness, M.~G. Bellemare,
  A.~Graves, M.~Riedmiller, A.~K. Fidjeland, G.~Ostrovski, {\em et~al.},
  ``Human-level control through deep reinforcement learning,'' {\em Nature},
  vol.~518, no.~7540, p.~529, 2015.

\bibitem{silver2016mastering}
D.~Silver, A.~Huang, C.~J. Maddison, A.~Guez, L.~Sifre, G.~Van Den~Driessche,
  J.~Schrittwieser, I.~Antonoglou, V.~Panneershelvam, M.~Lanctot, {\em et~al.},
  ``Mastering the game of go with deep neural networks and tree search,'' {\em
  nature}, vol.~529, no.~7587, p.~484, 2016.

\bibitem{kober2013reinforcement}
J.~Kober, J.~A. Bagnell, and J.~Peters, ``Reinforcement learning in robotics: A
  survey,'' {\em The International Journal of Robotics Research}, vol.~32,
  no.~11, pp.~1238--1274, 2013.

\bibitem{chebotar2017combining}
Y.~Chebotar, K.~Hausman, M.~Zhang, G.~Sukhatme, S.~Schaal, and S.~Levine,
  ``Combining model-based and model-free updates for trajectory-centric
  reinforcement learning,'' in {\em Proceedings of the 34th International
  Conference on Machine Learning-Volume 70}, pp.~703--711, JMLR. org, 2017.

\bibitem{DBLP:journals/corr/abs-1711-05253}
A.~Nagabandi, G.~Yang, T.~Asmar, G.~Kahn, S.~Levine, and R.~S. Fearing,
  ``Neural network dynamics models for control of under-actuated legged
  millirobots,'' {\em CoRR}, vol.~abs/1711.05253, 2017.

\bibitem{wahlstrom2015pixels}
N.~Wahlstr{\"o}m, T.~B. Sch{\"o}n, and M.~P. Deisenroth, ``From pixels to
  torques: Policy learning with deep dynamical models,'' {\em arXiv preprint
  arXiv:1502.02251}, 2015.

\bibitem{speyer2010primer}
J.~L. Speyer and D.~H. Jacobson, {\em Primer on optimal control theory},
  vol.~20.
\newblock Siam, 2010.

\bibitem{eren2017model}
U.~Eren, A.~Prach, B.~B. Ko{\c{c}}er, S.~V. Rakovi{\'c}, E.~Kayacan, and
  B.~A{\c{c}}{\i}kme{\c{s}}e, ``Model predictive control in aerospace systems:
  Current state and opportunities,'' {\em Journal of Guidance, Control, and
  Dynamics}, vol.~40, no.~7, pp.~1541--1566, 2017.

\bibitem{hedjar2013adaptive}
R.~Hedjar, ``Adaptive neural network model predictive control,'' {\em
  International Journal of Innovative Computing, Information and Control},
  vol.~9, no.~3, pp.~1245--1257, 2013.

\bibitem{chowdhary2013concurrent}
G.~Chowdhary, M.~M{\"u}hlegg, J.~P. How, and F.~Holzapfel, ``Concurrent
  learning adaptive model predictive control,'' in {\em Advances in Aerospace
  Guidance, Navigation and Control}, pp.~29--47, Springer, 2013.

\bibitem{argall2009survey}
B.~D. Argall, S.~Chernova, M.~Veloso, and B.~Browning, ``A survey of robot
  learning from demonstration,'' {\em Robotics and autonomous systems},
  vol.~57, no.~5, pp.~469--483, 2009.

\bibitem{schaal1997learning}
S.~Schaal, ``Learning from demonstration,'' in {\em Advances in neural
  information processing systems}, pp.~1040--1046, 1997.

\bibitem{li2011maximizing}
Z.~Li, D.~Glozman, D.~Milutinovic, and J.~Rosen, ``Maximizing dexterous
  workspace and optimal port placement of a multi-arm surgical robot,'' in {\em
  Robotics and Automation (ICRA), 2011 IEEE International Conference on},
  pp.~3394--3399, IEEE, 2011.

\bibitem{opencv_library}
G.~Bradski, ``{The OpenCV Library},'' {\em Dr. Dobb's Journal of Software
  Tools}, 2000.

\bibitem{kingma2014adam}
D.~P. Kingma and J.~Ba, ``Adam: A method for stochastic optimization,'' {\em
  arXiv preprint arXiv:1412.6980}, 2014.

\bibitem{Conti03}
F.~Conti, F.~Barbagli, R.~Balaniuk, M.~Halg, C.~Lu, D.~Morris, L.~Sentis,
  J.~Warren, O.~Khatib, and K.~Salisbury, ``The chai libraries,'' in {\em
  Proceedings of Eurohaptics 2003}, (Dublin, Ireland), pp.~496--500, 2003.

\bibitem{silva2009phantom}
A.~J. Silva, O.~A.~D. Ramirez, V.~P. Vega, and J.~P.~O. Oliver, ``Phantom omni
  haptic device: Kinematic and manipulability,'' in {\em Electronics, Robotics
  and Automotive Mechanics Conference, 2009. CERMA'09.}, pp.~193--198, IEEE,
  2009.

\end{thebibliography}

\end{document}